\documentclass[11pt]{article}
\usepackage{amsmath}
\usepackage{amsfonts}

\usepackage[preprint]{acl}

\usepackage{times}
\usepackage{latexsym}

\usepackage[T1]{fontenc}

\usepackage[utf8]{inputenc}

\usepackage{microtype}

\usepackage{inconsolata}
\usepackage{multirow}

\usepackage{graphicx}
\usepackage{float}
\usepackage{tabularx}
\usepackage{booktabs}
\usepackage{subcaption}
\usepackage{xcolor}

\title{Can Reasoning and Image Description Enhance Yes/No Visual Mathematical Understanding?}

\author{Guangcheng Song, Weichuan Wang\thanks{Corresponding authors.}, Linqi Song\footnotemark[1] \\
  City University of Hong Kong \\
  \texttt{\{guangsong5-c, weicwang2-c\}@my.cityu.edu.hk} \\
  \texttt{linqi.song@cityu.edu.hk}}

\begin{document}
\maketitle
\begin{abstract}
Large Vision-Language Models (VLMs) have achieved strong performance on general visual tasks, yet their ability to understand mathematical images remains limited. This paper investigates whether explicit reasoning and textual image descriptions can enhance VLMs' performance on binary yes/no visual mathematical questions. To achieve this, we curated a paired evaluation set to benchmark prominent open-source VLMs under varying prompt and input conditions. In particular, we leverage Chain-of-Thought (CoT) prompting and use textual descriptions, either replacing or augmenting visual inputs, to isolate and evaluate mathematical comprehension. Our results show that explicit reasoning improves performance, especially in paired consistency accuracy, while textual image descriptions do not consistently help and may degrade performance. These findings suggest that reasoning is important for stable visual mathematical judgement, whereas description-based intermediates require more careful design.
\end{abstract}

\section{Introduction}
\label{sec:introduction}

Large Vision-Language Models (VLMs) have become a central paradigm for connecting visual perception with natural language understanding. Representative models, such as Flamingo \cite{alayrac2022flamingo}, BLIP-2 \cite{li2023blip2}, LLaVA \cite{liu2023visual}, and InstructBLIP \cite{dai2023instructblip}, have shown strong image-text understanding and instruction-following abilities across diverse visual tasks. Recent evaluations further extend VLMs beyond natural images to structured visual inputs, including charts in ChartQA \cite{masry2022chartqa} and figures in FigureQA \cite{kahou2017figureqa}, where visual elements are organized by semantic, spatial, or symbolic relations. These advances make VLMs a promising interface for interpreting images that combine visual, textual, and symbolic information.

Mathematical images are important in mathematics education, where visual representations and spatial structures support conceptual understanding and problem solving \cite{SCHOENHERR2024100639,jintelligence11060127}. They also provide a valuable class of structured visual inputs for multimodal evaluation, as they typically combine geometric figures, coordinate plots, symbolic notations, numerical labels, and spatial relations. Although benchmarks such as MathVista \cite{lu2023mathvista} and MathVerse \cite{zhang2024mathverse} emphasize fine-grained diagrammatic configurations in visual mathematical reasoning, the extent to which current VLMs can exploit these configurations for reliable mathematical judgements remains unclear.

\begin{figure}
  \centering
  \includegraphics[width=\columnwidth]{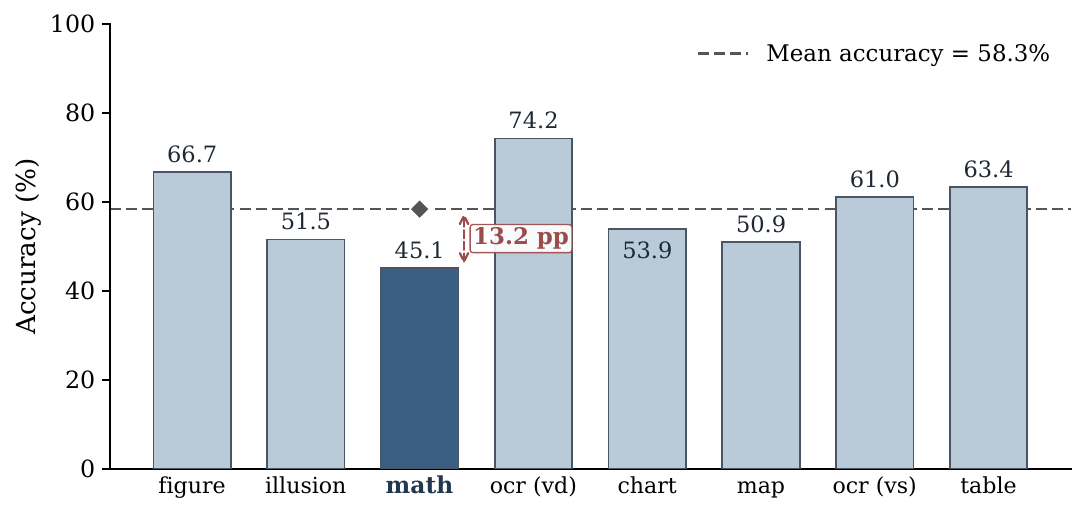}
  \caption{Accuracy of open-source VLMs across the eight static visual understanding categories in HallusionBench, averaged over 17 models from four model families. Detailed model-wise accuracies are reported in Appendix~\ref{app:appendix_hallusionbench}. The mathematical image category achieves 45.1\% accuracy, 13.2 percentage points (pp) below the category mean of 58.3\%.}
  \vspace{-4pt}
  \label{fig:result_1}
\end{figure}

To better characterize this challenge, we evaluate 17 open-source VLMs from four model families across the eight static visual understanding categories in HallusionBench \cite{guan2024hallusionbench}, whose questions are formulated as yes/no judgements. As illustrated in Figure~\ref{fig:result_1}, the mathematical image category yields an accuracy of only 45.1\%, falling 13.2 percentage points below the average performance of all static categories. This pronounced performance discrepancy indicates that mathematical image understanding remains a notable weakness of current VLMs, particularly when they must convert fine-grained diagrammatic configurations into reliable yes/no judgements.

The observed gap motivates the central question of this work: \textbf{\textit{Can reasoning and image description enhance yes/no visual mathematical understanding in VLMs?}} We address this question with an annotated paired evaluation set and a controlled study of reasoning and image description. For reasoning, we examine whether chain-of-thought prompting and thinking-oriented models improve VLMs beyond direct yes/no prediction. Our results show that explicit reasoning improves performance for most models, especially in paired consistency accuracy, suggesting its importance for stable mathematical judgement. For image description, inspired by caption-based methods for general visual question answering \cite{wu-etal-2019-generating, Hu_2023_ICCV, Guo_2023_CVPR}, we evaluate textual descriptions both as replacements for and augmentations to visual inputs. We find that textual descriptions do not consistently improve performance and may degrade it in some settings, indicating that naive description-based intermediates require more careful design for mathematical image understanding.

\section{Related Work}
\label{sec:related_work}

\paragraph{Large Vision-Language Models and Visual Mathematics}

Large Vision-Language Models (VLMs) connect visual encoders with large language models through image-text alignment, enabling models to process images with natural language prompts. Representative models, including BLIP-2 \cite{li2023blip2}, LLaVA \cite{liu2023visual}, and InstructBLIP \cite{dai2023instructblip}, have demonstrated strong image-text understanding ability. Recent open-source models, such as LLaVA-NeXT \cite{li2025llavanextinterleave}, Qwen3-VL \cite{bai2025qwen3vltechnicalreport}, and InternVL3.5 \cite{wang2025internvl35advancingopensourcemultimodal}, further extend these capabilities to interleaved multimodal inputs, long-context modeling, and improved reasoning. Alongside these developments, visual mathematical benchmarks have been proposed to evaluate VLMs on structured mathematical inputs. MathVista \cite{lu2023mathvista} and MathVerse \cite{zhang2024mathverse} evaluate visual mathematical reasoning and diagram use, while MATH-Vision \cite{wang2024mathvision} and DynaMath \cite{zou2024dynamath} focus on competition-level and dynamically generated visual math problems. These studies show that mathematical images remain challenging for VLMs, particularly when problems involve diagrams, symbols, numerical labels, and spatial relations.

\paragraph{Mathematical Reasoning and Captioning}

Recent methods improve multimodal mathematical reasoning by strengthening the connection between visual evidence and textual reasoning. MINT-CoT \cite{chen2026mintcot}, MathCanvas \cite{shi2025mathcanvasintrinsicvisualchainofthought}, and VLM-R$^3$ \cite{jiang2026vlmr} introduce visual or multimodal chain-of-thought mechanisms to better incorporate visual information during reasoning. Another line of work studies caption-based intermediate representations for general visual question answering. Question-relevant captioning \cite{wu-etal-2019-generating}, prompt-guided task-aware captioning \cite{Hu_2023_ICCV}, and image-to-text prompting with frozen language models \cite{Guo_2023_CVPR} convert visual inputs into textual forms to support answer prediction. Unlike these prior studies, our work provides a controlled study of whether explicit reasoning and textual image descriptions enhance yes/no mathematical image understanding in VLMs.

\section{Dataset Preparation}
\label{sec:dataset_preparation}

To evaluate VLMs on binary visual mathematical questions, we construct \textbf{MathYN}, a visual mathematical yes/no question answering dataset, from three public benchmarks: MathVista \cite{lu2023mathvista}, DynaMath \cite{zou2024dynamath}, and the mathematical category of HallusionBench. The construction process consists of three steps.

\noindent\textbf{Step 1: Question normalization}
Since the source benchmarks contain heterogeneous question formats, we select direct yes/no questions and binary-choice judgement questions, and convert them into a unified binary format. Each selected sample consists of an image, a question, and a yes/no answer. This step yields 267 candidate samples.

\noindent\textbf{Step 2: Mathematical filtering}
Question-type filtering still includes samples that mainly require real-world object recognition with shallow mathematical judgement, rather than mathematical image understanding. We therefore perform human annotation to filter mathematically relevant samples, following the instructions in Appendix~\ref{app:annotation_instruction}. Four volunteers, all graduate students or above in mathematics or computer science, independently label each sample as \textit{math} or \textit{not math}. To obtain a high-precision subset, we retain only samples unanimously labeled as \textit{math}.

\noindent\textbf{Step 3: Negated variants}
After annotation, 169 samples are retained, including 84 from MathVista (49.70\%), 51 from DynaMath (30.18\%), and 34 from HallusionBench-math (20.12\%). For each retained question, we construct a negated variant by reversing the semantic judgement while keeping the same image, forming paired yes/no questions with opposite answers. This results in the final MathYN dataset with 338 yes/no questions, as shown in Table~\ref{tab:mathyn_statistics}.

\begin{table}[t]
\centering
\small
\setlength{\tabcolsep}{2pt}
\begin{tabular}{lr@{\hspace{5pt}}lc}
\toprule
\multicolumn{2}{c}{Dataset} & \multicolumn{2}{c}{Agreement} \\
\cmidrule(lr){1-2} \cmidrule(lr){3-4}
Item & Number & Metric & Value \\
\midrule
Candidates & 267 & Pairwise Agreement & 0.9669 \\
Unanimous math & 169 & Fleiss' Kappa & 0.9261 \\
Negated variants & 169 & Gwet's AC1 & 0.9401 \\
Final questions & 338 & Unanimous Agreement & 0.9363 \\
\bottomrule
\end{tabular}
\caption{Statistics of the MathYN dataset and inter-annotator agreement for mathematical filtering.}
\vspace{-3pt}
\label{tab:mathyn_statistics}
\end{table}

\noindent\textbf{Annotation agreement.}
To assess annotation reliability, we compute the inter-annotator agreement across the 267 candidate samples. As shown in Table~\ref{tab:mathyn_statistics}, all metrics demonstrate strong consensus among the four annotators, confirming the suitability of this dataset for the mathematical domain.

\section{Experiments}
\label{sec:experiments}

\subsection{Experimental Setup}

\paragraph{Dataset and Models}
We evaluate models on MathYN, the dataset constructed in Section~\ref{sec:dataset_preparation}. The evaluated models are InternVL3.5-8B \cite{wang2025internvl35advancingopensourcemultimodal}, LLaMA3-LLaVA-NeXT-8B\footnote{\href{https://huggingface.co/llava-hf/llama3-llava-next-8b-hf}{llava-hf/llama3-llava-next-8b-hf}}, Qwen3-VL-8B \cite{bai2025qwen3vltechnicalreport}, and Qwen3.5-9B\footnote{\href{https://huggingface.co/Qwen/Qwen3.5-9B}{Qwen/Qwen3.5-9B}}.

\paragraph{Metrics}
We report accuracy and paired consistency accuracy. Let $\mathcal{D}=\{(x_i,y_i)\}_{i=1}^{N}$ be the full evaluation set, where $y_i \in \{\text{yes},\text{no}\}$ is the ground-truth answer and $\hat{y}_i$ is the model prediction. Accuracy is defined as: 
\[ \mathrm{Acc}=\frac{1}{N}\sum_{i=1}^{N}\mathbb{I}(\hat{y}_i=y_i). \]
For paired consistency accuracy, let $\mathcal{P}=\{(x_j^{o},y_j^{o},x_j^{n},y_j^{n})\}_{j=1}^{M}$ denote the original-negated question pairs. A pair is correct only if both questions are answered correctly:
\[ \mathrm{ConsAcc}=\frac{1}{M}\sum_{j=1}^{M} \mathbb{I}(\hat{y}_j^{o}=y_j^{o}) \mathbb{I}(\hat{y}_j^{n}=y_j^{n}). \] 
Here, $N=338$ and $M=169$. We extract the final yes/no decision from model outputs by regular expression matching; the extraction rule and its failure rate are given in Appendix~\ref{app:implementation_details}.

\subsection{Effect of Reasoning}
\label{sec:reasoning_experiment}

\begin{table}[t]
\centering
\small
\setlength{\tabcolsep}{2.5pt}
\begin{tabular}{lcc|cc}
\toprule
Model & Acc-D & Acc-C & Cons-D & Cons-C \\
\midrule
InternVL3.5-8B & 68.34 & \textbf{70.71} & 56.21 & \textbf{60.95} \\
LLaVA-NeXT-8B & \textbf{59.17} & 55.62 & \textbf{41.42} & 18.34 \\
Qwen3-VL-8B & 77.51 & \textbf{81.66} & 66.27 & \textbf{73.96} \\
Qwen3.5-9B & 80.18 & \textbf{81.36} & 69.23 & \textbf{74.56} \\
\bottomrule
\end{tabular}
\caption{Accuracy and paired consistency accuracy (\%) under direct and chain-of-thought prompting. D and C denote direct and chain-of-thought prompting, respectively. Bold indicates the better result between direct and chain-of-thought prompting for each model and metric.}
\vspace{-3pt}
\label{tab:reasoning_prompt}
\end{table}

\begin{table*}[t]
\centering
\small
\setlength{\tabcolsep}{3.5pt}
\begin{tabular}{llccc|ccc}
\toprule
Metric & Model & Img-D & Desc-D & Aug-D & Img-C & Desc-C & Aug-C \\
\midrule
\multirow{4}{*}{Acc.}
& InternVL3.5-8B & \textbf{68.34} & 63.61 & \textbf{68.34} & 70.71 & 71.89 & \textbf{73.08} \\
& LLaVA-NeXT-8B & \textbf{59.17} & 54.14 & 57.10 & \textbf{55.62} & 55.03 & 55.33 \\
& Qwen3-VL-8B & \textbf{77.51} & 68.93 & 71.89 & \textbf{81.66} & 74.26 & 75.44 \\
& Qwen3.5-9B & \textbf{80.18} & 74.85 & 78.70 & 81.36 & 82.25 & \textbf{83.14} \\
\midrule
\multirow{4}{*}{Cons.}
& InternVL3.5-8B & 56.21 & 52.66 & \textbf{57.99} & 60.95 & 62.72 & \textbf{65.09} \\
& LLaVA-NeXT-8B & \textbf{41.42} & 24.85 & 34.91 & 18.34 & 18.93 & \textbf{23.67} \\
& Qwen3-VL-8B & \textbf{66.27} & 55.62 & 59.17 & \textbf{73.96} & 63.31 & 65.68 \\
& Qwen3.5-9B & \textbf{69.23} & 62.13 & 68.64 & 74.56 & 75.74 & \textbf{79.29} \\
\bottomrule
\end{tabular}
\caption{Results (\%) under different input formats and prompting strategies. Img, Desc, and Aug denote image-question, description-question, and image-description-question formats, respectively. D and C denote direct and chain-of-thought prompting. Acc. and Cons. denote accuracy and paired consistency accuracy. Bold indicates the best result within each prompting group.}
\vspace{-6pt}
\label{tab:description_results}
\end{table*}

\paragraph{Chain-of-Thought Prompting}
To examine whether explicit reasoning improves yes/no mathematical image understanding, we compare two prompting settings: a direct prompt and a chain-of-thought prompt. The direct prompt asks the model to provide a binary yes/no answer, while the chain-of-thought prompt asks the model to reason step by step before giving the final answer. For each question, we provide the image and the corresponding question as input, extract the final yes/no prediction, and compare it with the ground truth. The detailed prompt templates are provided in Appendix~\ref{app:prompts}.

Table~\ref{tab:reasoning_prompt} reports accuracy and paired consistency accuracy under the two prompting settings. Chain-of-thought prompting consistently improves both metrics for three of the four models, with particularly clear gains in paired consistency accuracy. For example, Qwen3-VL-8B improves from 66.27\% to 73.96\%, and Qwen3.5-9B improves from 69.23\% to 74.56\%. These results suggest that explicit reasoning helps models make more stable judgements across paired yes/no mathematical questions. A decomposition of accuracy on original and negated questions is reported in Appendix~\ref{app:paired_decomposition}. LLaVA-NeXT-8B does not produce an explicit reasoning chain under this prompt and returns only the final judgement (Appendix~\ref{app:cot_response_length}).

\paragraph{Thinking-Oriented Models}

To further examine the effect of reasoning ability on mathematical image understanding, we compare instruct models with their thinking variants using the same direct-prompt template (Appendix~\ref{app:direct_prompt}). As shown in Table~\ref{tab:thinking_models}, thinking variants improve both accuracy and paired consistency accuracy in the two model families. The gains are especially clear for paired consistency accuracy: Qwen3-VL-8B improves from 66.27\% to 73.96\%, and Qwen3.5-9B improves from 69.23\% to 76.33\%. This suggests that thinking-oriented models are better at maintaining consistent judgements across original-negated question pairs.

\begin{table}[t]
\centering
\small
\setlength{\tabcolsep}{4pt}
\begin{tabular}{lcc}
\toprule
Model & Acc & Cons \\
\midrule
Qwen3-VL-8B & 77.51 & 66.27 \\
Qwen3-VL-8B-Thinking & \textbf{80.47} & \textbf{73.96} \\
\midrule
Qwen3.5-9B & 80.18 & 69.23 \\
Qwen3.5-9B-Thinking & \textbf{81.36} & \textbf{76.33} \\
\bottomrule
\end{tabular}
\caption{Accuracy and paired consistency accuracy (\%) of instruct and thinking models. Bold indicates the better result between the instruct and thinking variants within each model family.}
\vspace{-6pt}
\label{tab:thinking_models}
\end{table}

Together with the prompt-based comparison, these results indicate that reasoning is important for yes/no mathematical image understanding, especially for producing stable and consistent judgements across paired questions.

\subsection{Effect of Image Description}
\label{sec:description_experiment}

\paragraph{Setup}
We examine whether textual image descriptions can improve yes/no mathematical image understanding. For each image, each evaluated model first generates a description of its own, which is cached and reused in both the \textit{Desc} and \textit{Aug} settings (details in Appendix~\ref{app:implementation_details}). We then compare three input formats: \textit{Img} uses the original image and question, \textit{Desc} uses the generated description and question, and \textit{Aug} uses the image, description, and question together. Each format is evaluated with both direct and chain-of-thought prompts. The image description prompt and the answer prediction prompts are provided in Appendix~\ref{app:prompts}.

\paragraph{Results}
Table~\ref{tab:description_results} reports accuracy and paired consistency accuracy. Overall, description-only inputs often underperform the original image-question format, indicating that textual descriptions may lose information needed for mathematical image understanding. For example, under direct prompting, Qwen3-VL-8B drops from 77.51\% to 68.93\% in accuracy and from 66.27\% to 55.62\% in paired consistency accuracy. Adding descriptions can also introduce negative effects for some models; for instance, Qwen3-VL-8B decreases from 81.66\% to 75.44\% in accuracy under chain-of-thought prompting. While the augmented format benefits some models, such as InternVL3.5-8B and Qwen3.5-9B under chain-of-thought prompting, the gains are not consistent. These results suggest that textual descriptions may complement visual inputs in some cases, but can also introduce noise and should be used with caution for mathematical image understanding.

Our case analysis in Appendix~\ref{app:image_description_cases} suggests that textual descriptions of mathematical images can fail in two ways. First, they may introduce hallucinated or incorrect information, which misleads subsequent reasoning. Second, they may omit key visual details required for answering the question, such as geometric relations, numerical labels, or symbolic constraints. These issues help explain why description-only inputs underperform and why augmented inputs do not always yield consistent gains in our experiments.

\section{Conclusion}
\label{sec:conclusion}

We study VLMs on binary yes/no visual mathematical questions and construct a paired visual mathematical yes/no question answering dataset. Through evaluations under different prompting and input settings, we find that explicit reasoning improves model performance, particularly in paired consistency accuracy, suggesting its importance for stable mathematical judgement. In contrast, textual image descriptions do not consistently improve performance and can sometimes introduce negative effects. These findings highlight the importance of reasoning and the limitations of naive description-based intermediates for yes/no mathematical image understanding.

\section*{Limitations}

\noindent\textbf{\textit{Model Scope}}
Our experiments are strictly bounded within a designated subset of open-source Vision-Language Models characterized by moderate parameter scales. While the selected models represent a diverse cross-section of prominent model families and successfully demonstrate the critical role of reasoning capacity, the exact magnitude of this reasoning dependency may vary across alternative model scales.

\noindent\textbf{\textit{Evaluation Formats}}
This work focuses exclusively on binary affirmative and negative visual mathematical questions. This constrained setting facilitates a controlled evaluation framework utilizing paired original and negated queries. However, mathematical benchmarks encompass diverse answer formats, which may exhibit varying performance profiles under automated reasoning settings, suggesting that more fine-grained reasoning evaluations warrant future investigation.

\noindent\textbf{\textit{The Influence of Image Descriptions}}
This study evaluates a direct prompting strategy to utilize image descriptions, which unexpectedly induces a performance degradation. Nevertheless, highly capable models with sophisticated visual understanding may generate more detailed and precise descriptions that could ultimately augment performance within the mathematical domain, a phenomenon for which we observe partial empirical evidence in Table~\ref{tab:description_results}.

\section*{Ethical Statement}
Our investigation into image descriptions reveals that direct textual prompting can introduce performance degradation, highlighting an ethical concern regarding automated decision-making in high-stakes domains. When a system relies on intermediate textual descriptions of visual data, it runs the risk of generating hallucinated details that misrepresent structural, spatial, or geometric truths. Deploying such systems without explicit multimodal safeguards could result in catastrophic downstream failures in fields like automated engineering inspections or medical chart analysis. We emphasize that our findings serve as an ethical caution against the unvetted deployment of description-reliant workflows in safety-critical applications.


\bibliography{custom}
\clearpage

\appendix

\section{Detailed Results on HallusionBench}
\label{app:appendix_hallusionbench}

To support the motivation analysis in Figure~\ref{fig:result_1}, we evaluate 17 open-source VLMs from four model families on HallusionBench \cite{guan2024hallusionbench}. The evaluated models include variants from Qwen3-VL, Qwen3.5, InternVL3.5, and LLaVA-NeXT. HallusionBench organizes yes/no questions into Visual Dependent (VD) and Visual Supplement (VS) categories. VD includes figure, illusion, math, ocr, and video, while VS includes chart, map, ocr, and table.

Following its original evaluation protocol, we use an LLM-as-a-Judge setting to determine whether model responses match the ground-truth yes/no labels. We use DeepSeek-V3.2~\cite{liu2025deepseek} as the judge model and follow the same judge-based evaluation format. For each fine-grained category, we compute the accuracy of each model. The category-level averages used in Figure~\ref{fig:result_1} are obtained by directly taking the arithmetic mean of model accuracies within each category.

Figure~\ref{fig:result_1} focuses on static visual understanding categories. Therefore, we exclude the video category from the static-category average, since video questions additionally require temporal understanding. The reported static-category mean is computed over the remaining eight categories: figure, illusion, math, ocr (VD), chart, map, ocr (VS), and table. Based on the results in Table~\ref{tab:hallusionbench_finegrained}, these averages correspond to 45.10\% for the mathematical image category and 58.34\% across the eight static categories.

\begin{table*}[t]
  \centering
  \small
  \resizebox{\textwidth}{!}{
  \begin{tabular}{lccccc|cccc}
    \toprule
    \textbf{Model} &
    \multicolumn{5}{c|}{\textbf{VD (591)}} &
    \multicolumn{4}{c}{\textbf{VS (538)}} \\
    \cmidrule(lr){2-6}
    \cmidrule(lr){7-10}
    &
    \textbf{\begin{tabular}[c]{@{}c@{}}figure\\(80)\end{tabular}} &
    \textbf{\begin{tabular}[c]{@{}c@{}}illusion\\(144)\end{tabular}} &
    \textbf{\begin{tabular}[c]{@{}c@{}}math\\(108)\end{tabular}} &
    \textbf{\begin{tabular}[c]{@{}c@{}}ocr\\(89)\end{tabular}} &
    \textbf{\begin{tabular}[c]{@{}c@{}}video\\(170)\end{tabular}} &
    \textbf{\begin{tabular}[c]{@{}c@{}}chart\\(206)\end{tabular}} &
    \textbf{\begin{tabular}[c]{@{}c@{}}map\\(96)\end{tabular}} &
    \textbf{\begin{tabular}[c]{@{}c@{}}ocr\\(81)\end{tabular}} &
    \textbf{\begin{tabular}[c]{@{}c@{}}table\\(155)\end{tabular}} \\
    \midrule
    Qwen3-VL-2B  & 70.00 & 53.47 & 46.30 & 82.02 & 41.18 & 47.57 & 52.08 & 54.32 & 66.45 \\
    Qwen3-VL-4B  & 77.50 & 55.56 & 46.30 & 80.90 & 45.88 & 57.77 & 50.00 & 66.67 & 69.03 \\
    Qwen3-VL-8B  & 77.50 & 59.03 & 50.00 & 77.53 & 42.35 & 68.45 & 61.46 & 71.60 & 78.06 \\
    Qwen3-VL-32B & 76.25 & 57.64 & 58.33 & 79.78 & 44.71 & 74.76 & 59.38 & 72.84 & 77.42 \\
    \midrule
    Qwen3.5-0.8B & 56.25 & 46.53 & 42.59 & 76.40 & 37.06 & 45.15 & 46.88 & 48.15 & 57.42 \\
    Qwen3.5-2B   & 67.50 & 53.47 & 48.15 & 78.65 & 37.65 & 52.91 & 42.71 & 56.79 & 61.94 \\
    Qwen3.5-4B   & 73.75 & 61.11 & 58.33 & 84.27 & 51.18 & 66.02 & 48.96 & 74.07 & 75.48 \\
    Qwen3.5-9B   & 78.75 & 56.94 & 51.85 & 78.65 & 51.18 & 69.42 & 59.38 & 72.84 & 77.42 \\
    Qwen3.5-27B  & 78.75 & 57.64 & 62.96 & 83.15 & 56.47 & 76.70 & 68.75 & 79.01 & 78.06 \\
    \midrule
    InternVL3.5-1B  & 51.25 & 46.53 & 43.52 & 61.80 & 35.29 & 34.47 & 36.46 & 45.68 & 52.26 \\
    InternVL3.5-2B  & 65.00 & 52.08 & 41.67 & 70.79 & 37.65 & 43.20 & 44.79 & 59.26 & 60.00 \\
    InternVL3.5-4B  & 62.50 & 50.00 & 45.37 & 75.28 & 42.94 & 54.37 & 51.04 & 64.20 & 63.23 \\
    InternVL3.5-8B  & 62.50 & 54.86 & 41.67 & 78.65 & 40.00 & 55.83 & 55.21 & 61.73 & 61.94 \\
    InternVL3.5-14B & 67.50 & 52.08 & 49.07 & 74.16 & 43.53 & 57.28 & 52.08 & 70.37 & 65.81 \\
    \midrule
    LLaVA-v1.6-Vicuna-7B  & 57.50 & 39.58 & 23.15 & 60.67 & 30.00 & 29.13 & 42.71 & 48.15 & 40.00 \\
    LLaVA-v1.6-Vicuna-13B & 52.50 & 35.42 & 21.30 & 55.06 & 28.24 & 38.84 & 45.83 & 46.91 & 47.10 \\
    LLaMA3-LLaVA-NeXT-8B  & 58.75 & 43.75 & 36.11 & 64.04 & 31.18 & 44.66 & 47.92 & 44.44 & 45.81 \\
    \bottomrule
  \end{tabular}
  }
  \caption{
    Fine-grained accuracy (\%) of 17 open-source VLMs on HallusionBench. VD and VS denote Visual Dependent and Visual Supplement categories, respectively. Figure~\ref{fig:result_1} is computed by averaging model accuracies within each static category and excluding the video category.
  }
  \label{tab:hallusionbench_finegrained}
\end{table*}

\section{Annotation Instruction}
\label{app:annotation_instruction}

This section provides the annotation instruction used for mathematical filtering in Section~\ref{sec:dataset_preparation}. The goal is to identify samples that require mathematical image understanding, rather than samples that can be answered mainly through general object recognition or commonsense visual understanding.

Each annotator is asked to assign one of the following two labels to each candidate sample:

\begin{itemize}
    \item \textbf{\textit{math}}: The sample should be labeled as \textit{math} if answering the question requires interpreting mathematical visual content in the image, such as geometric figures, coordinate plots, numerical labels, symbolic notations, mathematical diagrams, or spatial relations.
    \item \textbf{\textit{not math}}: The sample should be labeled as \textit{not math} if answering the question mainly depends on recognizing real-world objects, reading non-mathematical text, or applying commonsense visual understanding, even if the question contains simple numerical or comparative expressions.
\end{itemize}

\begin{figure*}[p]
    \centering
    \begin{minipage}[t]{0.47\textwidth}
        \centering
        \includegraphics[width=\linewidth]{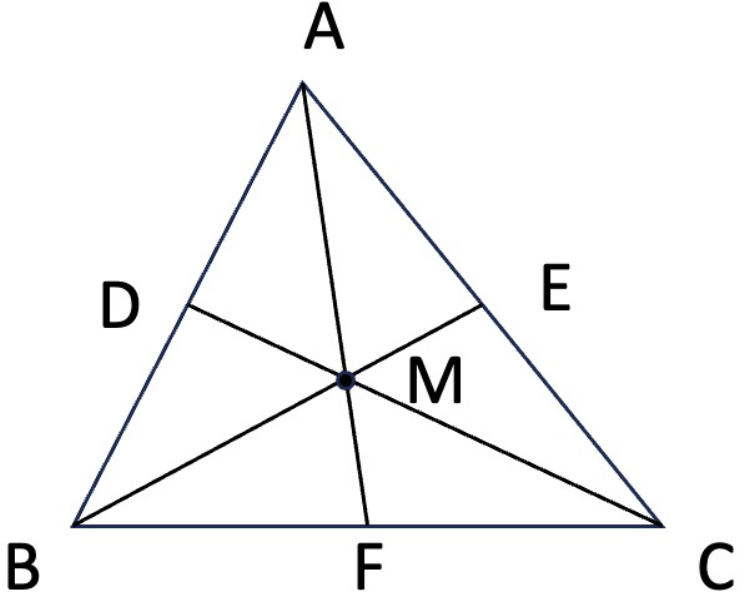}
        \vspace{1mm}
        \small (a) \textit{math}
    \end{minipage}
    \hfill
    \begin{minipage}[t]{0.47\textwidth}
        \centering
        \includegraphics[width=\linewidth]{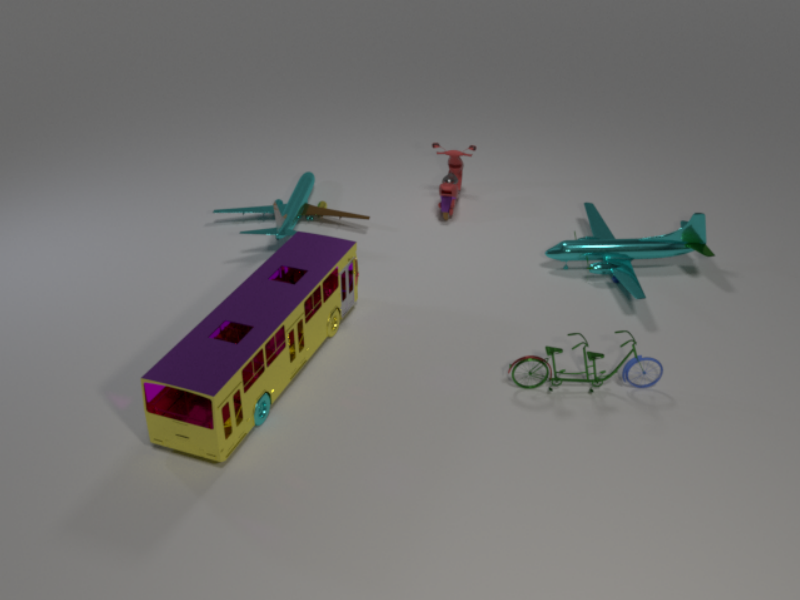}
        \vspace{1mm}
        \small (b) \textit{not math}
    \end{minipage}
    \caption{Examples of images labeled as \textit{math} and \textit{not math} during mathematical filtering.}
    \label{fig:annotation_examples}
\end{figure*}

Annotators label a sample as \textit{math} only when answering it requires mathematical visual evidence; otherwise, it is labeled as \textit{not math}. Examples are shown in Figure~\ref{fig:annotation_examples}. Each sample is annotated by four annotators, and only unanimously labeled \textit{math} samples are retained.

\section{Prompts}
\label{app:prompts}

This section lists the prompt templates used in our experiments. The placeholder \texttt{\{question\}} denotes the input question, and \texttt{\{description\}} denotes the generated image description.

\subsection{Direct Prompt}
\label{app:direct_prompt}

For the original image-question setting, the model receives the image and the following text prompt:

\begin{quote}
Answer the question based on the provided image and question. Answer directly with yes or no. Your response MUST be exactly in this format: \{Judgement: yes\} or \{Judgement: no\}.

Question: \texttt{\{question\}}
\end{quote}

\subsection{Chain-of-Thought Prompt}

For the chain-of-thought setting, the model receives the image and the following text prompt:

\begin{quote}
Answer the question based on the provided image and question. First reason carefully about the image and question. You may think step by step internally, but only output the final answer in this format: \{Judgement: yes\} or \{Judgement: no\}.

Question: \texttt{\{question\}}
\end{quote}

\subsection{Image Description Prompt}

To generate textual image descriptions, we use the following prompt:

\begin{quote}
Describe this image in detail. Only use visible information. Do not infer the hidden question or solve it.
\end{quote}

\subsection{Description-Based Question Prompt}

For both the description-question and image-description-question settings, we prepend the generated image description to the question prompt. The two settings use the same textual prompt; the only difference is that the image-description-question setting additionally provides the original image as visual input.

\paragraph{Direct.}

\begin{quote}
Image Description:

\texttt{\{description\}}

Answer the question based on the provided image and question. Answer directly with yes or no. Your response MUST be exactly in this format: \{Judgement: yes\} or \{Judgement: no\}.

Question: \texttt{\{question\}}
\end{quote}

\paragraph{Chain-of-thought.}

\begin{quote}
Image Description:

\texttt{\{description\}}

Answer the question based on the provided image and question. First reason carefully about the image and question. You may think step by step internally, but only output the final answer in this format: \{Judgement: yes\} or \{Judgement: no\}.

Question: \texttt{\{question\}}
\end{quote}

\section{Image Description Cases}
\label{app:image_description_cases}

\begin{figure*}[p]
    \centering
    \begin{minipage}[t]{0.32\textwidth}
        \centering
        \vspace{0pt}
        \includegraphics[width=\linewidth]{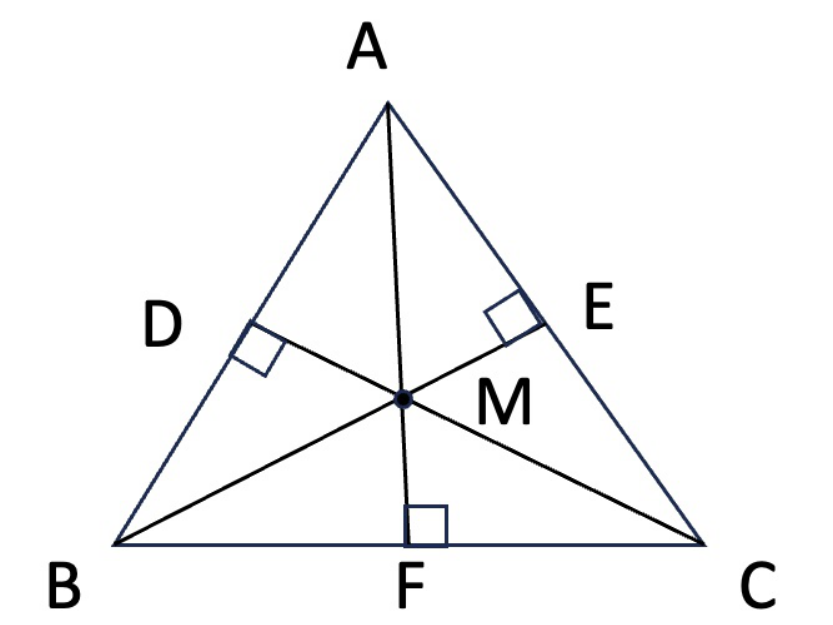}
    \end{minipage}
    \hfill
    \begin{minipage}[t]{0.64\textwidth}
        \vspace{0pt}
        \footnotesize
        \raggedright

        \textbf{Question:} If $BE \perp AC$, $CD \perp AB$, and $AF \perp BC$, then is $M$ the orthocenter of triangle $ABC$?

        \vspace{1mm}
        \textbf{Generated description:}
        {\ttfamily
        This is a geometric diagram of a triangle with several lines and points drawn inside it.\\
        - The main figure is a triangle labeled with vertices A, B, and C.\\
        - There are three points labeled D, E, and F, which lie on the sides of the triangle:\\
        - Point D is on side AB.\\
        - Point E is on side AC.\\
        - Point F is on side BC.\\
        - Three line segments are drawn from each vertex to the opposite side, connecting to points D, E, and F respectively:\\
        - Segment \textcolor{red}{AD} connects vertex A to point D on side AB.\\
        - Segment BE connects vertex B to point E on side AC.\\
        - Segment \textcolor{red}{CF} connects vertex C to point F on side BC.\\
        - \textcolor{red}{These three line segments (AD, BE, and CF) intersect at a single point inside the triangle}, labeled M.\\
        - At each of the points D, E, and F, there is a small square symbol drawn at the intersection of the respective line segment and the side of the triangle.\\
        This symbol indicates a right angle (90 degrees).\\
        - The square at D indicates \textcolor{red}{that segment AD is perpendicular to side AB}.\\
        - The square at E indicates that segment BE is perpendicular to side AC.\\
        - The square at F indicates that \textcolor{red}{segment CF is perpendicular to side BC}.\\
        - The lines AD, BE, and CF are drawn as thick, dark lines, while the sides of the triangle ABC are drawn as thinner, dark lines.\\
        - The point M is marked with a small, filled black dot.
        }

        \vspace{1mm}
        \textbf{Issue:} The red text marks hallucinated or incorrect visual information in the generated description, which can mislead the subsequent reasoning process.
    \end{minipage}
    \caption{A case where the generated image description introduces hallucinated or incorrect information.}
    \label{fig:desc_case_hallucination}
\end{figure*}

\begin{figure*}[p]
    \centering
    \begin{minipage}[t]{0.32\textwidth}
        \centering
        \vspace{0pt}
        \includegraphics[width=\linewidth]{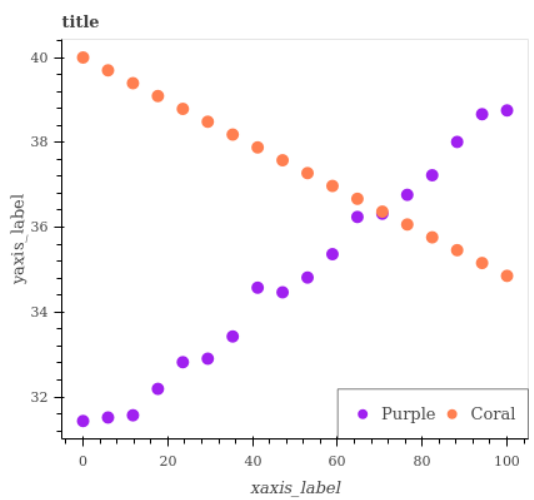}
    \end{minipage}
    \hfill
    \begin{minipage}[t]{0.64\textwidth}
        \vspace{0pt}
        \footnotesize
        \raggedright

        \textbf{Question:} Is Coral the roughest?

        \vspace{1mm}
        \textbf{Generated description:}
        {\ttfamily
        This is a scatter plot with two distinct data series plotted against a grid.\\
        - **Title**: The plot has a title labeled "title" at the top center.\\
        - **Axes**:\\
        \hspace*{2mm}- The horizontal axis (x-axis) is labeled "xaxis\_label" and ranges from 0 to 100.\\
        \hspace*{2mm}- The vertical axis (y-axis) is labeled "yaxis\_label" and ranges from 32 to 40.\\
        - **Data Series**: There are two sets of data points, distinguished by color and a legend in the bottom right corner.\\
        \hspace*{2mm}- **Purple Series**: Represented by purple circular markers. This series starts at an x-value near 0 with a y-value of approximately 31.5, and the points generally trend upward as the x-value increases, ending near an x-value of 100 with a y-value of approximately 38.5.\\
        \hspace*{2mm}- **Coral Series**: Represented by coral-colored circular markers. This series starts at an x-value near 0 with a y-value of approximately 40, and the points generally trend downward as the x-value increases, ending near an x-value of 100 with a y-value of approximately 34.5.\\
        - **Legend**: A legend box is located in the bottom right corner. It contains two entries: a purple circle labeled "Purple" and a coral circle labeled "Coral".\\
        - **Grid**: A light gray grid is visible, with lines corresponding to the tick marks on both axes.\\
        - **Plot Area**: The data points are plotted within the defined axes, with no lines connecting the points.
        }

        \vspace{1mm}
        \textbf{Issue:} Although the description is not necessarily incorrect, it omits key visual details required to answer the question.
    \end{minipage}
    \caption{A case where the generated image description omits question-relevant visual information.}
    \label{fig:desc_case_missing_info}
\end{figure*}

This section provides representative cases to illustrate why textual image descriptions do not consistently improve mathematical image understanding. We observe two common failure patterns: descriptions may introduce hallucinated or incorrect information, as shown in Figure~\ref{fig:desc_case_hallucination}, and they may omit visual details that are necessary for answering the question, as shown in Figure~\ref{fig:desc_case_missing_info}.

\section{Paired Accuracy Decomposition}
\label{app:paired_decomposition}

Table~\ref{tab:paired_decomposition} decomposes the results in Table~\ref{tab:reasoning_prompt} into accuracy on original questions (Acc-O) and negated questions (Acc-N). The consistency gains from chain-of-thought prompting are not uniform across the two sides: InternVL3.5-8B improves on both, Qwen3-VL-8B and Qwen3.5-9B improve mainly on negated questions, and LLaVA-NeXT-8B declines on both.

\begin{table*}[t]
\centering
\small
\setlength{\tabcolsep}{6pt}
\begin{tabular}{l cc cc cc cc}
\toprule
& \multicolumn{2}{c}{Acc} & \multicolumn{2}{c}{Acc-O} & \multicolumn{2}{c}{Acc-N} & \multicolumn{2}{c}{Cons} \\
\cmidrule(lr){2-3} \cmidrule(lr){4-5} \cmidrule(lr){6-7} \cmidrule(lr){8-9}
Model & D & C & D & C & D & C & D & C \\
\midrule
InternVL3.5-8B & 68.34 & \textbf{70.71} & 71.01 & \textbf{74.56} & 65.68 & \textbf{66.86} & 56.21 & \textbf{60.95} \\
LLaVA-NeXT-8B & \textbf{59.17} & 55.62 & \textbf{59.76} & 57.99 & \textbf{58.58} & 53.25 & \textbf{41.42} & 18.34 \\
Qwen3-VL-8B & 77.51 & \textbf{81.66} & \textbf{84.02} & 84.02 & 71.01 & \textbf{79.29} & 66.27 & \textbf{73.96} \\
Qwen3.5-9B & 80.18 & \textbf{81.36} & \textbf{87.57} & 83.43 & 72.78 & \textbf{79.29} & 69.23 & \textbf{74.56} \\
\bottomrule
\end{tabular}
\caption{Accuracy (\%) on all questions (Acc), original questions (Acc-O), and negated questions (Acc-N), together with paired consistency accuracy (Cons), under direct (D) and chain-of-thought (C) prompting. Bold indicates the better result between D and C for each model and metric.}
\label{tab:paired_decomposition}
\end{table*}

\section{Response Length under Chain-of-Thought}
\label{app:cot_response_length}

Table~\ref{tab:cot_response_length} reports raw response length and the judgement-only rate on all 338 MathYN questions under the chain-of-thought prompt in Appendix~\ref{app:prompts}. Response length is the whitespace-delimited word count of the complete response, including the final judgement, and is reported as the median with the interquartile range (25th--75th percentiles). The judgement-only rate is the proportion of outputs that contain only \texttt{\{Judgement: yes\}} or \texttt{\{Judgement: no\}} and no additional text. Under this prompt, LLaVA-NeXT-8B has a median length of 2.0 words and a judgement-only rate of 100\%, whereas the other models produce substantially longer responses. LLaVA-NeXT-8B therefore does not produce an explicit reasoning chain here, and its paired consistency accuracy of 18.34\% reflects performance under this prompting condition.

\begin{table}[h]
\centering
\small
\setlength{\tabcolsep}{5pt}
\begin{tabular}{lcc}
\toprule
Model & Median [IQR] & Judgement-only \\
\midrule
InternVL3.5-8B & 97.5 [64.0, 137.5] & 0.00\% \\
LLaVA-NeXT-8B & 2.0 [2.0, 2.0] & 100.00\% \\
Qwen3-VL-8B & 186.0 [127.0, 265.8] & 8.58\% \\
Qwen3.5-9B & 187.5 [119.2, 297.0] & 0.00\% \\
\bottomrule
\end{tabular}
\caption{Raw response length in words (median and interquartile range) and judgement-only rate under chain-of-thought prompting, measured on all 338 MathYN questions.}
\label{tab:cot_response_length}
\end{table}

\section{Implementation Details}
\label{app:implementation_details}

All model inference in our main experiments is conducted using the vLLM framework. We set the decoding temperature to 0.0 for generation.

We extract the final yes/no decision with a case-insensitive regular expression for \texttt{\{Judgement: yes\}} or \texttt{\{Judgement: no\}}, and fall back to an unambiguous yes or no when this formatted judgement is absent. The parsing success rate is 99.55\%. Outputs that cannot be parsed are counted as incorrect.

For image descriptions (Section~\ref{sec:description_experiment}), each evaluated model generates one description per image using the prompt in Appendix~\ref{app:prompts}, with a maximum length of 2,048 tokens. The description is generated once, cached, and reused unchanged in both the description-question and image-description-question settings for that model.

\section{AI Assistant}
I acknowledge the use of artificial intelligence tools for copyediting, grammar checking, and structural polishing during the revision phase of this text. These tools were used strictly to enhance readability, flow, and professional tone, while the core ideas, arguments, and content remain entirely our own.

\clearpage

\end{document}